\documentclass[10pt,twocolumn,letterpaper]{article}

\usepackage{cvpr}
\usepackage{times}
\usepackage{epsfig}
\usepackage{graphicx}
\usepackage{amsmath}
\usepackage{amssymb}
\usepackage{caption}
\usepackage{booktabs}
\usepackage{multirow}
\usepackage{mathtools}

\usepackage[pagebackref=true,breaklinks=true,letterpaper=true,colorlinks,bookmarks=false]{hyperref}

\cvprfinalcopy 

\def\cvprPaperID{10609} 

\ifcvprfinal\fi

\newcommand{\E}{\mathbb{E}}
\newcommand{\KL}{{\rm KL}}

\begin{document}

\title{Flow Contrastive Estimation of Energy-Based Models}

\author{Ruiqi Gao$^{1}$, Erik Nijkamp$^{1}$ , Diederik P. Kingma$^{2}$, Zhen Xu$^{2}$, Andrew M. Dai$^{2}$, Ying Nian Wu$^{1}$\\
$^{1}$ UCLA, $^{2}$ Google \\ {\tt\small $\{$ruiqigao, enijkamp$\}$@ucla.edu, $\{$durk, zhenxu, adai$\}$@google.com, ywu@stat.ucla.edu}\\
}


\maketitle
\begin{abstract}
   This paper studies a training method to jointly estimate an energy-based model and a flow-based model, in which the two models are iteratively updated based on a shared adversarial value function. This joint training method has the following traits. (1) The update of the energy-based model is based on noise contrastive estimation, with the flow model serving as a strong noise distribution. (2) The update of the flow model approximately minimizes the Jensen-Shannon divergence between the flow model and the data distribution. (3) Unlike generative adversarial networks (GAN) which estimates an implicit probability distribution defined by a generator model, our method estimates two explicit probabilistic distributions on the data. Using the proposed method we demonstrate a significant improvement on the synthesis quality of the flow model, and show the effectiveness of unsupervised feature learning by the learned energy-based model. Furthermore, the proposed training method can be easily adapted to semi-supervised learning. We achieve competitive results to the state-of-the-art semi-supervised learning methods.
\end{abstract}

\section{Introduction}
Recently, \emph{flow-based models} (henceforth simply called \emph{flow models}) have gained popularity as a type of deep generative model \cite{dinh2014nice,dinh2016density, kingma2018Glow,grathwohl2018ffjord,behrmann2018invertible, kumar2019videoflow, tran2019discrete,durkan2019neural} and for use in variational inference \cite{kingma2013auto,rezende2015variational, kingma2016improved}. 

Flow models have two properties that set them apart from other types of deep generative models: (1) they allow for efficient evaluation of the density function, and (2) they allow for efficient sampling from the model. Efficient evaluation of the log-density allows flow models to be directly optimized towards the log-likelihood objective, unlike variational autoencoders (VAEs) \cite{kingma2013auto, rezende2014stochastic}, which are optimized towards a \emph{bound} on the log-likelihood, and generative adversarial networks (GANs) \cite{goodfellow2014generative}. Auto-regressive models~\cite{graves2013generating,oord2016wavenet,salimans2017pixelcnn}, on the other hand, are (in principle) inefficient to sample from, since synthesis requires computation that is proportional to the dimensionality of the data.

These properties of efficient density evaluation and efficient sampling are typically viewed as advantageous. However, they have a potential downside: these properties also acts as \emph{assumptions} on the true data distribution that they are trying to model. By choosing a flow model, one is making the assumption that the true data distribution is one that is in principle simple to sample from, and is computationally efficient to normalize. In addition, flow models assume that the data is generated by a finite sequence of invertible functions. If these assumptions do not hold, flow-based models can result in a poor fit. 

On the other end of the spectrum of deep generative models lies the family of energy-based models (EBMs) \cite{lecun2006tutorial,ngiam2011learning,kim2016deep,zhao2016energy,xie2016theory,gao2018learning,kumar2019maximum,nijkamp2019learning,du2019implicit,finn2016connection,goyal2017variational,grathwohl2019your,desjardins2011tracking}.
Energy-based models define an unnormalized density that is the exponential of the negative \emph{energy function}. The energy function is directly defined as a (learned) scalar function of the input, and is often parameterized by a neural network, such as a convolutional network \cite{lecun1998gradient,krizhevsky2012imagenet}.
Evaluation of the density function for a given data point involves calculating a normalizing constant, which requires an intractable integral. Sampling from EBMs is expensive and requires approximation as well, such as computationally expensive Markov Chain Monte Carlo (MCMC) sampling.  
EBMs, therefore, do not make any of the two assumptions above: they do not assume that the density of data is easily normalized, and they do not assume efficient synthesis. Moreover, they do not constrain the data distribution by invertible functions.  


Contrasting an EBM with a flow model, the former is on the side of representation where different layers represent features of different complexities, whereas the latter is on the side of learned computation, where each layer, or each transformation, is like a step in the computation. The EBM is like an objective function or a target distribution whereas the flow model is like a finite step iterative algorithm or a learned sampler.  Borrowing language from reinforcement learning \cite{finn2016connection}, the flow model is like an actor whereas the EBM is like a critic or an evaluator. The EBM can be simpler and more flexible in form than the flow model which is highly constrained, and thus the EBM may capture  the modes of the data distribution more accurately than the flow model. 
In contrast, the flow model is capable of direct generation via ancestral sampling,  which is sorely lacking in an EBM. It may thus be desirable to train the two models jointly, combining the tractability of flow model and the flexibility of EBM. This is the goal of this paper.

Our joint training method is inspired by the noise contrastive estimation (NCE) of \cite{gutmann2010noise}, where an EBM is learned discriminatively by classifying the real data and the data generated by a noise model. In NCE, the noise model must have an explicit normalized density function. Moreover, it is desirable for the noise distribution to be close to the data distribution for accurate estimation of the EBM. However, the noise distribution can be far away from the data distribution. The flow model can potentially transform or transport the noise distribution to a distribution closer to the data distribution. With the advent of strong flow-based generative models \cite{dinh2014nice, dinh2016density, kingma2018Glow}, it is natural to recruit the flow model as the contrast distribution for noise contrastive estimation of the EBM.

However, even with the flow-based model pre-trained by maximum likelihood estimation (MLE) on the data distribution, it may still not be strong enough as a contrast distribution, in the sense that the synthesized examples generated by the pre-trained flow model may still be distinguished from the real examples by a classifier based on an EBM.  Thus, we want the flow model to be a stronger contrast or a stronger training opponent for EBM. To achieve this goal, we can simply use the same objective function of NCE, which is the log-likelihood of the logistic regression for classification. While NCE updates the EBM by maximizing this objective function, we can also update the flow model by minimizing the same objective function to make the classification task harder for the EBM. Such update of flow model combines MLE and variational approximation, and helps correct the over-dispersion of MLE.  If the EBM is close to the data distribution, this amounts to minimizing the Jensen-Shannon divergence (JSD) \cite{goodfellow2014generative} between the data distribution and the flow model.  In this sense, the learning scheme relates closely to GANs \cite{goodfellow2014generative}. However, unlike GANs, which learns a generator model that defines an implicit probability density
 function via a low-dimensional latent vector, our method learns two probabilistic models with explicit probability densities (a normalized one and an unnormalized one). 
 
The contributions of our paper are as follows. We explore a parameter estimation method that couples estimation of an EBM and a flow model using a shared objective function. It improves NCE with a flow-transformed noise distribution, and it modifies MLE of the flow model to approximate JSD minimization, and helps correct the over-dispersion of MLE. Experiments on 2D synthetic data show that the learned EBM achieves accurate density estimation with a much simpler network structure than the flow model. On real image datasets, we demonstrate a significant improvement on the synthesis quality of the flow model, and the effectiveness of unsupervised feature learning by the energy-based model. Furthermore, we show that the proposed method can be easily adapted to semi-supervised learning, achieving performance comparable to state-of-the-art semi-supervised methods. 

\section{Related work}
For learning the energy-based model by MLE, the main difficulty lies in drawing fair samples from the current model. A prominent approximation of MLE is the contrastive divergence (CD) \cite{hinton2002training} framework, requiring MCMC initialized from the data distribution. CD has been generalized to persistent CD \cite{tieleman2008training}, and has more recently been generalized to modified CD \cite{gao2018learning}, adversarial CD \cite{kim2016deep, dai2017calibrating, han2018divergence} with modern CNN structure. \cite{nijkamp2019learning, du2019implicit} scale up sampling-based methods to large image datasets with white noise as the starting point of sampling. However, these sampling based methods may still have difficulty traversing different modes of the learned model, which may result in biased model, and may take a long time to converge. Another variant is to  An advantage of noise contrastive estimation (NCE), and our adaptive version of it, is that it avoids MCMC sampling in estimation of the energy-based model, by turning the estimation problem into a classification problem.


Generalizing from \cite{tu2007learning}, \cite{jin2017introspective, lazarow2017introspective, lee2018wasserstein} developed an introspective parameter estimation method, where the EBM is discriminatively learned and composed of a sequence of discriminative models obtained through the learning process. Another line of work is to estimate the parameters of EBM by score matching \cite{hyvarinen2005estimation,zhai2016deep,saremi2018deep,song2019generative}. \cite{zhai2019adversarial,finn2016connection} connects GAN to the estimation of EBM. 

NCE and it variants has gained popularity in natural language processing (NLP) \cite{he2016training, oualil2017batch, baltescu2014pragmatic, bose2018adversarial}. \cite{mnih2012fast, mnih2013learning} applied NCE to log-bilinear models and in \cite{vaswani2013decoding} NCE is applied to neural probabilistic language models. NCE shows effectiveness in typical NLP tasks such as word embeddings \cite{mikolov2013distributed} and order embeddings \cite{vendrov2015order}. 

In the context of inverse reinforcement learning, \cite{levine2013guided} proposes a guided policy search method, and \cite{finn2016connection} connects it to GAN.  Our method is closely related to this method, where the energy function can be viewed as the cost function, and the flow model can be viewed as the unrolled policy. 

\section{Learning method}

\subsection{Energy-based model}
\label{sect: ebm}
Let $x$ be the input variable, such as an image. We use $p_\theta(x)$ to denote a model's probability density function of $x$ with parameter $\theta$. The energy-based model (EBM) is defined as follows:
\begin{equation}
    p_\theta(x) = \frac{1}{Z(\theta)}\exp[f_\theta(x)],
    \label{model: ebm}
\end{equation}
where $f_\theta(x)$ is defined by a bottom-up convolutional neural network whose parameters are denoted by $\theta$. The normalizing constant $Z(\theta) = \int \exp[ f_\theta(x) ] dx$ is intractable to compute exactly for high-dimensional $x$.

\subsubsection{Maximum likelihood estimation}
The energy-based model in eqn. \ref{model: ebm} can be estimated from unlabeled data by maximum likelihood estimation (MLE). Suppose we observe training examples $\{x_i, i = 1,...,n\}$ from unknown true distribution $p_{\rm data}(x)$. We can view this dataset as forming empirical data distribution, and thus expectation with respect to $p_{\rm data}(x)$ can be approximated by averaging over the training examples. In MLE, we seek to maximize the log-likelihood function
\begin{equation}
L(\theta) = \frac{1}{n} \sum_{i=1}^n \log p_\theta(x_i).
\end{equation}
Maximizing the log-likelihood function is equivalent to minimizing the Kullback-Leibler divergence $\KL (p_{\rm data} || p_\theta)$ for large $n$. Its gradient can be written as:
\begin{equation}
- \frac{\partial}{\partial \theta} \KL(p_{\rm data}||p_\theta) = \E_{p_{\rm data}}\left[\frac{\partial}{\partial \theta} f_\theta(x)\right] - \E_{p_\theta}\left[\frac{\partial}{\partial \theta}f_\theta(x)\right],
\end{equation}
which is the difference between the expectations of the gradient of $f_\theta (x)$ under $p_{\rm data}$ and $p_\theta$ respectively. The expectations can be approximated by averaging over the observed examples and synthesized samples generated from the current model $p_\theta(x)$ respectively. The difficulty lies in the fact that sampling from $p_\theta(x)$ requires MCMC such as Hamiltonian monte carlo or Langevin dynamics \cite{girolami2011riemann, zhu1998grade}, which may take a long time to converge, especially on high dimensional and multi-modal space such as image space.

The MLE of $p_\theta(x)$ seeks to cover all the models of $p_{\rm data}(x)$. Given the flexibility of model form of $f_\theta(x)$, the MLE of $p_\theta(x)$ has the chance to approximate $p_{\rm data}(x)$ reasonably well. 

\subsubsection{Noise contrastive estimation}
\label{section: nce}
Noise contrastive estimation (NCE) \cite{gutmann2010noise} can be used to learn the EBM, by including the normalizing constant as another learnable parameter. Specifically, for an energy-based model $p_{\theta}(x) = \frac{1}{Z(\theta)}\exp[f_{\theta}(x)]$, we define $p_{\theta}(x) = \exp[f_{\theta}(x) - c]$, where $c = \log Z(\theta)$. $c$ is now treated as a free parameter, and is included into $\theta$. Suppose we observe training examples  $\{x_i, i = 1,...,n\}$, and we have generated examples $\{\tilde{x}_i, i = 1,...,n\}$ from a noise distribution $q(x)$. Then $\theta$ can be estimated by maximizing the following objective function: 
\begin{equation}
	\resizebox{.88\linewidth}{!}{$J(\theta) = \E_{p_{\rm data}}\left[ \log \frac{p_\theta(x)}{p_\theta(x) +  q(x)}\right] + \E_q\left[\log\frac{q(x)}{p_\theta(x) + q(x)}\right]$},
	\label{eqn: nce}
\end{equation}
which transforms estimation of EBM into a classification problem. 

The objective function connects to logistic regression in supervised learning in the following sense. Suppose for each training or generated examples we assign a binary class label $y$: $y = 1$ if $x$ is from training dataset and $y = 0$ if $x$ is generated from $q(x)$. In logistic regression, the posterior probabilities of classes given the data $x$ are estimated. As the data distribution $p_{\rm data}(x)$ is unknown, the class-conditional probability $p(\cdot|y = 1)$ is modeled with $p_\theta(x)$. And $p(\cdot|y=0)$ is modeled by $q(x)$. Suppose we assume equal probabilities for the two class labels, i.e., $p(y=1) = p(y=0) = 0.5 $. Then we obtain the posterior probabilities:
\begin{equation}
p_\theta(y=1|x) = \frac{p_\theta(x)}{p_\theta(x) + q(x)} \coloneqq u(x, \theta).
\end{equation}
The class-labels $y$ are Bernoulli-distributed, so that the log-likelihood of the parameter $\theta$ becomes 
\begin{equation}
	\begin{split}
		l(\theta) = \sum_{i=1}^n\log u(x_i;\theta) + \sum_{i=1}^n \log(1 - u(\tilde{x}_i; \theta)),
	\end{split}
\end{equation}
which is, up to a factor of $1/n$, an approximation of eqn. \ref{eqn: nce}.

The choice of the noise distribution $q(x)$ is a design issue. Generally speaking, we expect $q(x)$ to satisfy the following: (1) analytically tractable expression of normalized density; (2) easy to draw samples from; (3) close to data distribution. In practice, (3) is important for learning a model over high dimensional data. If $q(x)$ is not close to the data distribution, the classification problem would be too easy and would not require $p_\theta$ to learn much about the modality of the data. 

\subsection{Flow-based model}
\label{sect: flow}
A flow model is of the form 
 \begin{equation} 
 x = g_\alpha(z); \; z \sim q_0(z),
 \end{equation}
where $q_0$ is a known noise distribution. $g_\alpha$ is a composition of a sequence of invertible transformations where the log-determinants of the Jacobians of the transformations can be explicitly obtained.  $\alpha$ denotes the parameters. Let $q_\alpha(x)$ be the probability density of the model given a datapoint $x$ with parameter $\alpha$. Then under the \emph{change of variables} $q_\alpha(x)$ can be expressed as
\begin{equation}
q_\alpha(x) = q_0(g_\alpha^{-1}(x))|\det(\partial g_\alpha^{-1}(x)/\partial x)|. 
\end{equation}

More specifically, suppose $g_\alpha$ is composed of a sequence of transformations $g_\alpha = g_{\alpha_1} \circ \cdots \circ g_{\alpha_m}$. The relation between $z$ and $x$ can be written as $z \leftrightarrow h_1 \leftrightarrow \cdots \leftrightarrow h_{m-1}\leftrightarrow x$. And thus we have 
\begin{equation}
	q_\alpha(x) = q_0(g_\alpha^{-1}(x)) \Pi_{i=1}^m |\det(\partial h_{i-1} / \partial h_i)|,
\end{equation}
where we define $z \coloneqq h_0$ and $x \coloneqq h_m$ for conciseness. With carefully designed transformations, as explored in flow-based methods, the determinant of the Jacobian matrix $(\partial h_{i-1} / \partial h_i)$ can be incredibly simple to compute. The key idea is to choose transformations whose Jacobian is a triangle matrix, so that the determinant becomes 
\begin{equation}
	|\det(\partial h_{i-1} / \partial h_i)| = \Pi|{\rm diag}(\partial h_{i-1} / \partial h_{i})|. 
\end{equation}

The following are the two scenarios for estimating $q_\alpha$:

(1) Generative modeling by MLE \cite{dinh2014nice,dinh2016density, kingma2018Glow,grathwohl2018ffjord,behrmann2018invertible, kumar2019videoflow, tran2019discrete}, based on $\min_\alpha {\rm KL}(p_{\rm data} \| q_\alpha)$, where again $\E_{p_{\rm data}}$ can be approximated by average over observed examples. 

(2) Variational approximation to an unnormalized target density $p$ \cite{kingma2013auto,rezende2015variational, kingma2016improved,kingma2014efficient, khemakhem2019variational}, based on $\min_\alpha {\rm KL}(q_\alpha\|p)$, where 
\begin{equation}
\begin{split} 
	& \resizebox{.78\linewidth}{!}{$\KL (q_\alpha\|p) = \E_{q_\alpha}[\log q_\alpha(x)] - \E_{q_\alpha}[\log p(x)]$}\\
    & \resizebox{.88\linewidth}{!}{$= \E_{z} [\log q_0(z) - \log |{\rm det}(g_\alpha'(z))|] - \E_{q_\alpha}[\log p(x)]. $} 
\end{split}
\end{equation} 
${\rm KL}(q_\alpha\|p)$ is the difference between energy and entropy, i.e., we want $q_\alpha$ to have low energy but high entropy. ${\rm KL}(q_\alpha\|p)$ can be calculated without inversion of $g_\alpha$. 

When $q_\alpha$ appears on the right of KL-divergence, as in (1), it is forced to cover most of the modes of $p_{\rm data}$, When $q_\alpha$ appears on the left of KL-divergence, as in (2), it tends to chase the major modes of $p$ while ignoring the minor modes \cite{murphy2012machine,fox2012tutorial}. As shown in the following section, our proposed method learns a flow model by combining (1) and (2). 

\subsection{Flow Contrastive Estimation}

A natural improvement to NCE is to transform the noise so that the resulting distribution is closer to the data distribution. This is exactly what the flow model achieves. That is, a flow model transform a known noise distribution $q_0(z)$ by a composition of a sequence of invertible transformations $g_\alpha(\cdot)$. It also fulfills (1) and (2) of the requirements of NCE. However, in practice, we find that a pre-trained $q_\alpha(x)$, such as learned by MLE, is not strong enough for learning an EBM $p_\theta(x)$ because the synthesized data from the MLE of $q_\alpha(x)$ can still be easily distinguished from the real data by an EBM. Thus, we propose to iteratively train the EBM and flow model, in which case the flow model is adaptively adjusted to become a stronger contrast distribution or a stronger training opponent for EBM. This is achieved by a parameter estimation scheme similar to GAN, where $p_\theta(x)$ and $q_\alpha(x)$ play a minimax game with a unified value function: $ \min_{\alpha}\max_{\theta} V(\theta, \alpha)$, 
\begin{equation}
\begin{split}
	 V(\theta, \alpha) &=  \E_{p_{\rm data}}  \left[ \log \frac{p_\theta(x)}{p_\theta(x) +  q_\alpha(x)}\right] \\
	  &+  \E_{z}\left[\log \frac{q_\alpha(g_\alpha(z))}{p_\theta(g_\alpha(z)) + q_\alpha(g_\alpha(z))}\right],
\end{split}
\label{eqn: value}
\end{equation}
  where $\E_{p_{\rm data}}$ is approximated by averaging over observed samples $\{x_i, i = 1,...,n\}$, while $\E_z$ is approximated by averaging over negative samples $\{\tilde{x}_i, i = 1,...,n\}$ drawn from $q_\alpha(x)$, with $z_i \sim q_0(z)$ independently for $i = 1, ..., n$. In the experiments, we choose Glow \cite{kingma2018Glow} as the flow-based model. The algorithm can either start from a randomly initialized Glow model or a pre-trained one by MLE. Here we assume equal prior probabilities for observed samples and negative samples. It can be easily modified to the situation where we assign a higher prior probability to the negative samples, given the fact we have access to infinite amount of free negative samples. 

The objective function can be interpreted from the following perspectives:

(1) Noise contrastive estimation for EBM. The update of $\theta$ can be seen as noise contrastive estimation of $p_\theta(x)$, but with a flow-transformed noise distribution $q_\alpha(x)$ which is adaptively updated. The training is essentially a logistic regression. However, unlike regular logistic regression for classification, for each $x_i$ or $\tilde{x}_i$, we must include $\log q_\alpha(x_i)$ or $\log q_\alpha(\tilde{x}_i)$ as an example-dependent bias term. This forces $p_\theta(x)$ to replicate $q_\alpha(x)$ in addition to distinguishing between $p_{\rm data}(x)$ and $q_\alpha(x)$, so that $p_\theta(x_i)$ is in general larger than $q_\alpha(x_i)$, and $p_\theta(\tilde{x}_i)$ is in general smaller than $q_\alpha(\tilde{x}_i)$. 

(2) Minimization of Jensen-Shannon divergence for the flow model. If $p_\theta(x)$ is close to the data distribution, then the update of $\alpha$ is approximately minimizing the Jensen-Shannon divergence between the flow model $q_\alpha$ and data distribution $p_{\rm data}$: 
\begin{equation}
\begin{split}
{\rm JSD}(q_\alpha \| p_{\rm data})&={\rm KL}(p_{\rm data}  \| (p_{\rm data} + q_\alpha)/2) \\
&+ {\rm KL}(q_\alpha \| (p_{\rm data} + q_\alpha)/2).
\end{split}
\end{equation}
Its gradient w.r.t. $\alpha$ equals the gradient of $-\E_{p_{\rm data}}[\log((p_\theta + q_\alpha)/2)] + {\rm KL}(q_\alpha \| (p_\theta + q_\alpha)/2)$. The gradient of the first term resembles MLE, which forces $q_\alpha$ to cover the modes of data distribution, and tends to lead to an over-dispersed model, which is also pointed out in \cite{kingma2018Glow}. The gradient of the second term is similar to reverse Kullback-Leibler divergence between $q_\alpha$ and $p_\theta$, or variational approximation of $p_\theta$ by $q_\alpha$, which forces $q_\alpha$ to chase the modes of $p_\theta$ \cite{murphy2012machine,fox2012tutorial}. This may help correct the over-dispersion of MLE, and combines the two scenarios of estimating the flow-based model $q_\alpha$ as described in section~\ref{sect: flow}.  

(3) Connection with GAN. 
Our parameter estimation scheme is closely related to GAN. In GAN, the discriminator $D$ and generator $G$ play a minimax game: $\min_G \max_D V(G, D)$,
\begin{equation}
    V(G, D) = \E_{p_{\rm data}} \left[ \log D(x)\right]
    + \E_z\left[ \log(1 - D(G(z_i)))\right].
\end{equation}
 The discriminator $D(x)$ is learning the probability ratio 
 	 $p_{\rm data}(x)/(p_{\rm data}(x) + p_G(x))$,
 which is about the difference between $p_{\rm data}$ and $p_G$ \cite{finn2016connection}. In the end, if the generator $G$ learns to perfectly replicate $p_{\rm data}$, then the discriminator $D$ ends up with a random guess. However, in our method, the  ratio is explicitly modeled by $p_\theta$ and $q_\alpha$. $p_\theta$ must contain all the learned knowledge in $q_\alpha$, in addition to the difference between $p_{\rm data}$ and $q_\alpha$. In the end, we learn two explicit probability distributions $p_\theta$ and $q_\alpha$ as approximations to $p_{\rm data}$.  
 
Henceforth we simply refer to the proposed method as flow constrastive estimation, or FCE. 

\subsection{Semi-supervised learning}
\label{sect: semi}
A class-conditional energy-based model can be transformed into a discriminative model in the following sense. Suppose there are $K$ categories $k = 1,..., K$, and the model learns a distinct density $p_{\theta_k}(x)$ for each $k$. The networks $f_{\theta_k}(x)$ for $k = 1,..., K$ may share common lower layers, but with different top layers. Let $\rho_k$ be the prior probability of category $k$, for $k = 1,..., K$. Then the posterior probability for classifying $x$ to the category $k$ is a softmax multi-class classifier
\begin{equation}
     P(k|x) = \frac{\exp(f_{\theta_k}(x) + b_k)}{\sum_{l=1}^K\exp(f_{\theta_l}(x) + b_l)},
     \label{model: cls}
\end{equation}
where $b_k =   \log(\rho_k) - \log Z(\theta_k)$. 


Given this correspondence, we can modify FCE to do semi-supervised learning. Specifically, assume $\{(x_i, y_i), i=1,...,m\}$ are observed examples with labels known, and $\{x_i, i = m+1,..., m+n\}$ are observed unlabeled examples. For each category $k$, we can assume that class-conditional EBM is in the form
\begin{equation}
p_{\theta_k}(x) = \frac{1}{Z(\theta_k)} \exp[f_{\theta_k}(x)] = \exp[f_{\theta_k}(x) - c_k],
\end{equation}  
where $f_{\theta_k}(x)$ share all the weights except for the top layer. And we assume equal prior probability for each category. Let $\theta$ denotes all the parameters from class-conditional EBMs $\{\theta_k, k = 1,..., K\}$. For labeled examples, we can maximize the conditional posterior probability of label $y$, given $x$ and the fact that $x$ is an observed example (instead of a generated example from $q_\alpha$). By Bayes rule, this leads to maximizing the following objective function over $\theta$:
\begin{equation}
\begin{split}
	L_{\rm label}(\theta)& = \E_{p_{\rm data}(x, y)}\left[\log p_\theta(y | x, y \in \{1, ..., K\}) \right] \\
	&= \E_{p_{\rm data}(x, y)}\left[ \log\frac{p_{\theta_y}(x)}{\sum_{k=1}^K p_{\theta_k}(x)} \right],
\end{split}
\label{eqn: label}
\end{equation}
which is similar to a classifier in the form. 

For unlabeled examples, the probability can be defined by an unconditional EBM, which is in the form of a mixture model:
\begin{equation}
	p_\theta(x) =\sum_{i=1}^K p_\theta(x|y=k) p(y=k) = \frac{1}{K} \sum_{i=1}^K p_{\theta_k}(x),
\end{equation}
Together with the generated examples from $q_\alpha(x)$, we can define the same value function $V(\theta, \alpha)$ as eqn. \ref{eqn: value} for the unlabeled examples. The joint estimation algorithm alternate the following two steps: (1) update $\theta$ by $\max_\theta L_{\rm label}(\theta) + V(\theta, \alpha)$; (2) update $\alpha$ by $\min_\alpha V(\theta, \alpha)$. Due to the flexibility of EBM, $f_{\theta_k}(x)$ can be defined by any existing state-of-the-art network structures designed for semi-supervised learning. 
\section{Experiments}

For FCE, we adaptively adjust the numbers of updates for EBM and Glow: we first update EBM for a few iterations until the classification accuracy is above $0.5$, and then we update Glow until the classification accuracy is below $0.5$. We use \emph{Adam} \cite{kingma2014adam} with learning rate $\alpha=0.0003$ for the EBM and \emph{Adamax} \cite{kingma2014adam} with learning rate $\alpha=0.00001$ for the Glow model. Code and more results can be found at {\small \url{http://www.stat.ucla.edu/~ruiqigao/fce/main.html}}

\subsection{Density estimation on 2D synthetic data}
\begin{figure}
\begin{center}
   \setlength{\tabcolsep}{2.5pt}
        \begin{tabular}{cccc}
        {\footnotesize Data} & {\footnotesize Glow-MLE} & {\footnotesize Glow-FCE} & {\footnotesize EBM-FCE}\\
             \includegraphics[width=.07\textwidth]{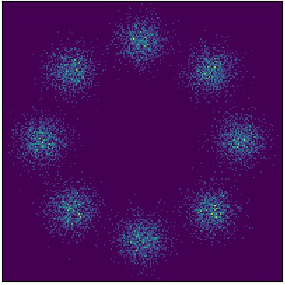} & \includegraphics[width=.07\textwidth]{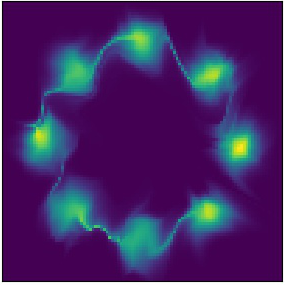} &
             \includegraphics[width=.07\textwidth]{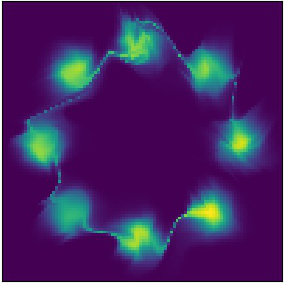} & \includegraphics[width=.07\textwidth]{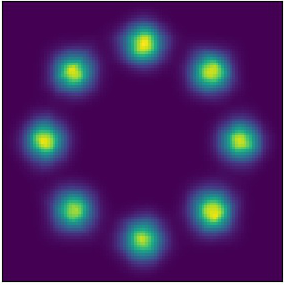}\\
             \includegraphics[width=.07\textwidth]{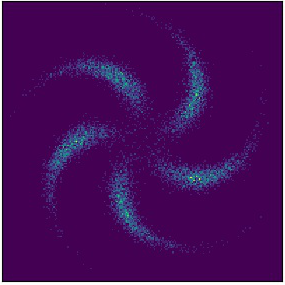} & \includegraphics[width=.07\textwidth]{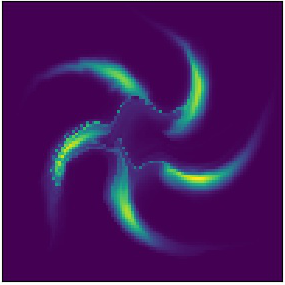} &
             \includegraphics[width=.07\textwidth]{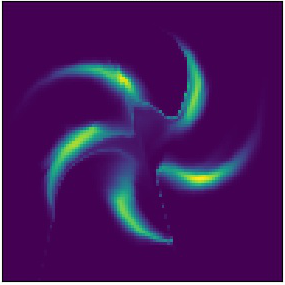} & \includegraphics[width=.07\textwidth]{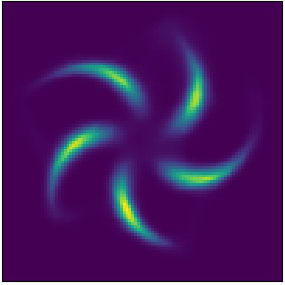} \\
             \includegraphics[width=.07\textwidth]{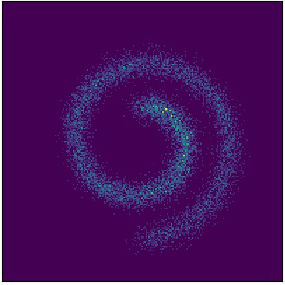} & \includegraphics[width=.07\textwidth]{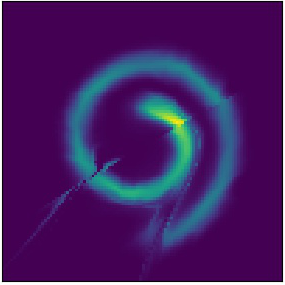} &
             \includegraphics[width=.07\textwidth]{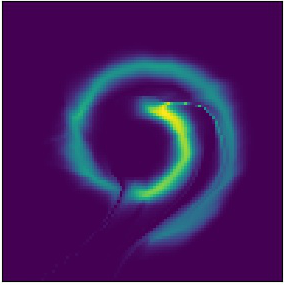} & \includegraphics[width=.07\textwidth]{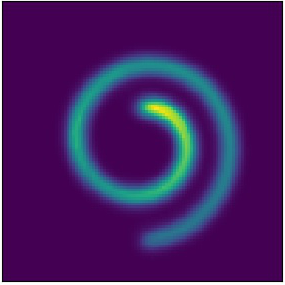}
        \end{tabular}\\
  \end{center}
  \caption{Comparison of trained EBM and Glow models on 2-dimensional data distributions.}
  \label{fig: 2d}
\end{figure}

Figure \ref{fig: 2d} demonstrates the results of FCE on several 2D distributions, where FCE starts from a randomly initialized Glow. The learned EBM can fit multi-modal distributions accurately, and forms a better fit than Glow learned by either FCE or MLE. Notably, the EBM is defined by a much simpler network structure than Glow: for Glow we use $10$ affine coupling layers, which amount to $30$ fully-connected layers, while the energy-based model is defined by a $4$-layer fully-connected network with the same width as Glow.  Another interesting finding is that the EBM can fit the distributions well, even if the flow model is not a perfect contrastive distribution.

\begin{figure}
	\begin{center}
	\includegraphics[width=.38\textwidth]{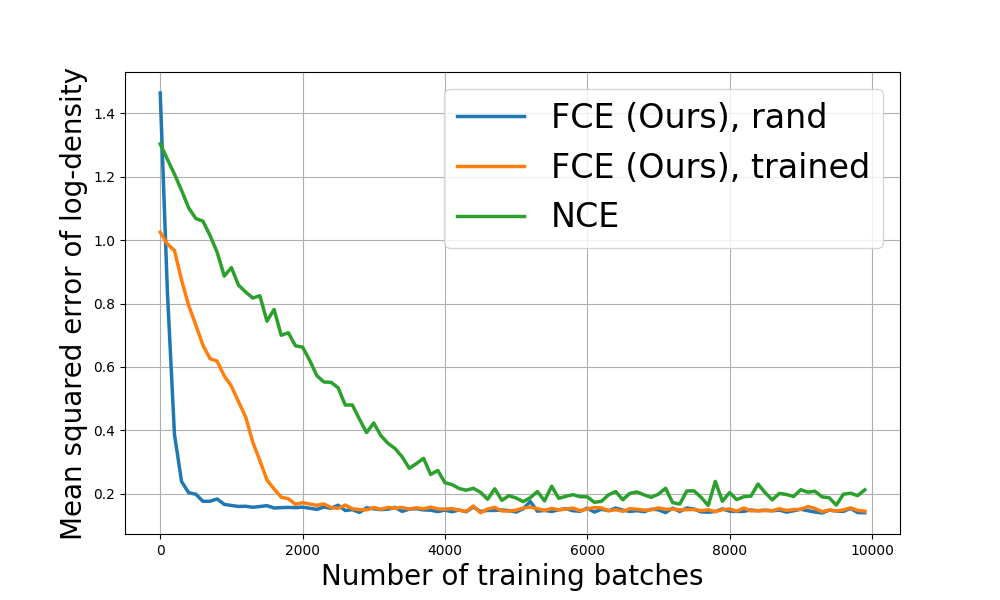}
\caption{Density estimation accuracy in 2D examples of a mixture of 8 Gaussian distributions.}
\label{fig: 2d_mse} 
	\end{center}
\end{figure}

For the distribution depicted in the first row of Figure~\ref{fig: 2d}, which is a mixture of eight Gaussian distributions, we can compare the estimated densities by the learned models with the ground truth densities. Figure \ref{fig: 2d_mse} shows the mean squared error of the estimated log-density over numbers of training iterations of EBMs. We show the results of FCE either starting from a randomly initialized Glow ('rand') or a Glow model pre-trained by MLE ('trained'), and compare with NCE with a Gaussian noise distribution. FCE starting from a randomly initialized Glow converges in fewer iterations. Both settings of FCE achieve a lower error rate than NCE.

\subsection{Learning on real image datasets}
\begin{figure*}
\begin{center}
\setlength{\tabcolsep}{8pt}
\begin{tabular}{ccc}
  \includegraphics[width=.24\textwidth]{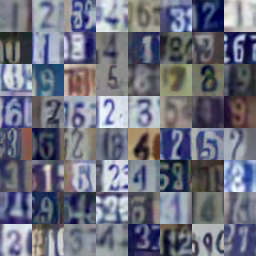}   & \includegraphics[width=.24\textwidth]{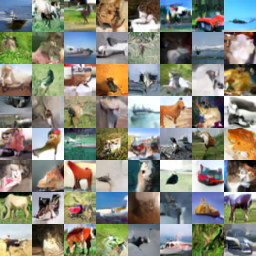} & 
  \includegraphics[width=.24\textwidth]{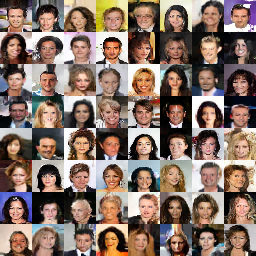}
\end{tabular}
\caption{Synthesized examples from the Glow model learned by FCE. From left to right panels are from SVHN, CIFAR-10 and CelebA datasets, respectively. The image size is $32 \times 32$.}
\label{fig:syn}
\end{center}
\end{figure*}
We conduct experiments on the Street View House Numbers (SVHN) \cite{netzer2011reading}, CIFAR-10 \cite{krizhevsky2009learning} and CelebA \cite{liu2015faceattributes} datasets. We resized the CelebA images to $32 \times 32$ pixels, and used $20,000$ images as a test set. We initialize FCE with a pre-trained Glow model, trained by MLE, for the sake of efficiency. We again emphasize the simplicity of the EBM model structure compared to Glow. See Appendix for detailed model architectures. For Glow, depth per level \cite{kingma2018Glow} is set as $8$, $16$, $32$ for SVHN, CelebA and CIFAR-10 respectively. Figure \ref{fig:syn} depicts synthesized examples from learned Glow models. To evaluate the fidelity of synthesized examples, Table \ref{tab: fid} summarizes the Fr\'echet Inception Distance (FID) \cite{heusel2017gans} of the synthesized examples computed with the Inception V3 \cite{szegedy2016rethinking} classifier. The fidelity is significantly improved compared to Glow trained by MLE (see Appendix for qualitative comparisons), and is competitive to the other generative models. In Table \ref{tab: llh}, we report the average negative log-likelihood (bits per dimension) on the testing sets. The log-likelihood of the learned EBM is based on the estimated normalizing constant (i.e., a parameter of the model) and should be taken with a grain of salt. 
For the learned Glow model, the log-likelihood of the Glow model estimated with FCE is slightly lower than the log-likelihood of the Glow model trained with MLE.

\begin{table}
\centering
 \caption{FID scores for generated samples. For our method, we evaluate generative samples from the learned Glow model.}
 \label{tab: fid}
 \footnotesize
  \begin{tabular}{lccc}
    \toprule
     Method   & SVHN & CIFAR-10 & CelebA \\
        \midrule
     VAE \cite{kingma2013auto}       &57.25 & 78.41& 38.76 \\
     DCGAN \cite{radford2015unsupervised} &   21.40 & 37.70 & 12.50 \\
     Glow \cite{kingma2018Glow} & 41.70 &  45.99 & 23.32\\
     FCE (Ours) & {\bf 20.19}  & {\bf 37.30} & {\bf 12.21} \\
        \bottomrule
    \end{tabular}
\end{table}

\begin{table}
\centering
 \caption{Bits per dimension on testing data. $^\dagger$ indicates that the log-likelihood is computed based on models with estimated normalizing constant, and should be taken with a grain of salt.}
 \label{tab: llh} 
 \setlength{\tabcolsep}{4pt}
 \footnotesize
  \begin{tabular}{lccc}
    \toprule
        Model & SVHN & CIFAR-10 & CelebA \\
        \midrule
        Glow-MLE &2.17    &  3.35 & 3.49 \\
        Glow-FCE (Ours) & 2.25& 3.45 & 3.54\\
        EBM-FCE (Ours) & $^\dagger$2.15  & $^\dagger$3.27 & $^\dagger$3.40 \\
        \bottomrule
    \end{tabular}
\end{table}


\subsection{Unsupervised feature learning}

\begin{figure}
\begin{center}
	\includegraphics[width=.38\textwidth]{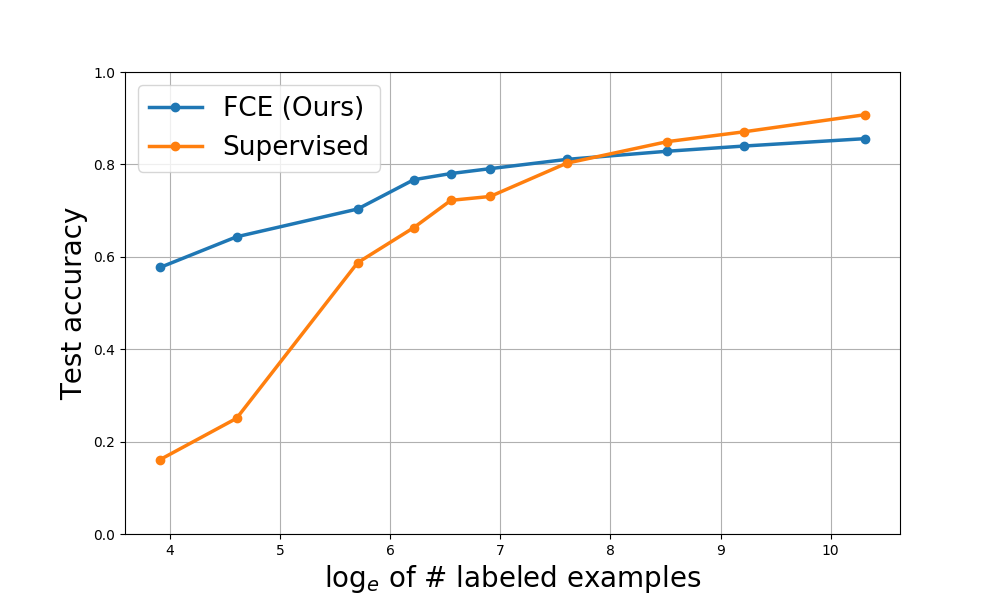}
    \captionof{figure}{SVHN test-set classification accuracy as a function of number of labeled examples. The features from top layer feature maps are extracted and a linear classifier is learned on the extracted features.}
    \label{fig: cls_acc}
\end{center}
\end{figure}

To further explore the EBM learned with FCE, we perform unsupervised feature learning with features from a learned EBM. Specifically, we first conduct FCE on the entire training set of SVHN in an unsupervised way. Then, we extract the top layer feature maps from the learned EBM, and train a linear classifier on top of the extracted features using only a subset of the training images and their corresponding labels.  Figure \ref{fig: cls_acc} shows the classification accuracy as a function of the number of labeled examples. Meanwhile, we compare our method with a supervised model with the same model structure as the EBM, and is trained only on the same subset of labeled examples each time. We observe that FCE outperforms the supervised model when the number of labeled examples is small (less than $2000$). 
 
Next we try to combine features from multiple layers together. Specifically, following the same procedure outlined in \cite{radford2015unsupervised}, the features from the top three convolutional layers are max pooled and concatenated to form a $14,336$-dimensional vector of feature. A regularized L2-SVM is then trained on these features with a subset of training examples and the corresponding labels. Table \ref{tabl: feature} summarizes the results of using $1,000$, $2,000$ and $4,000$ labeled examples from the training set. At the top part of the table, we compare with methods that estimate an EBM or a discriminative model coupled with a generator network. At the middle part of the table, we compare with methods that learn an EBM with contrastive divergence (CD) and modified versions of CD. For fair comparison, we use the same model structure for the EBMs or discriminative models used in all the methods. The results indicate that FCE outperforms these methods in terms of the effectiveness of learned features.  

\begin{table}
\centering
 \caption{Test set classification error of L2-SVM classifier trained on the concatenated features learned from SVHN. DDGM stands for Deep Directed Generative Models. For fair comparison, all the energy-based models or discriminative models are trained with the same model structure. }
 \footnotesize
  \begin{tabular}{lccc}
    \toprule
     \multirow{2}{*}{Method}   & \multicolumn{3}{c}{\# of labeled data} \\
     & $1000$ & $2000$ & $4000$\\
        \midrule
     WGAN \cite{wasserstein}       &43.15 & 38.00& 32.56\\
     WGAN-GP \cite{gulrajani2017improved} & 40.12 & 32.24 & 30.63 \\
     DDGM \cite{kim2016deep} &44.99& 34.26& 27.44\\
     DCGAN \cite{radford2015unsupervised} &   38.59 & 32.51 & 29.37\\
     SN-GAN \cite{miyato2018spectral} &40.82 & 31.24 & 28.69 \\
     MMD-GAN-rep \cite{wang2018improving} & 36.74 & 29.12 & 25.23 \\
     \midrule
     Persistent CD \cite{tieleman2008training} & 45.74 & 39.47 & 34.18 \\
     One-step CD \cite{hinton2002training} & 44.38 & 35.87 & 30.45 \\
     Multigrid sampling \cite{gao2018learning} & 30.23 & 26.54 & 22.83 \\
     \midrule
     FCE (Ours) & {\bf 27.07} & {\bf 24.12} & {\bf 22.05}\\
        \bottomrule
    \end{tabular}
    \label{tabl: feature}
\end{table}

\subsection{Semi-supervised learning}
\begin{figure*}
\begin{center}
	\includegraphics[width=.8\textwidth]{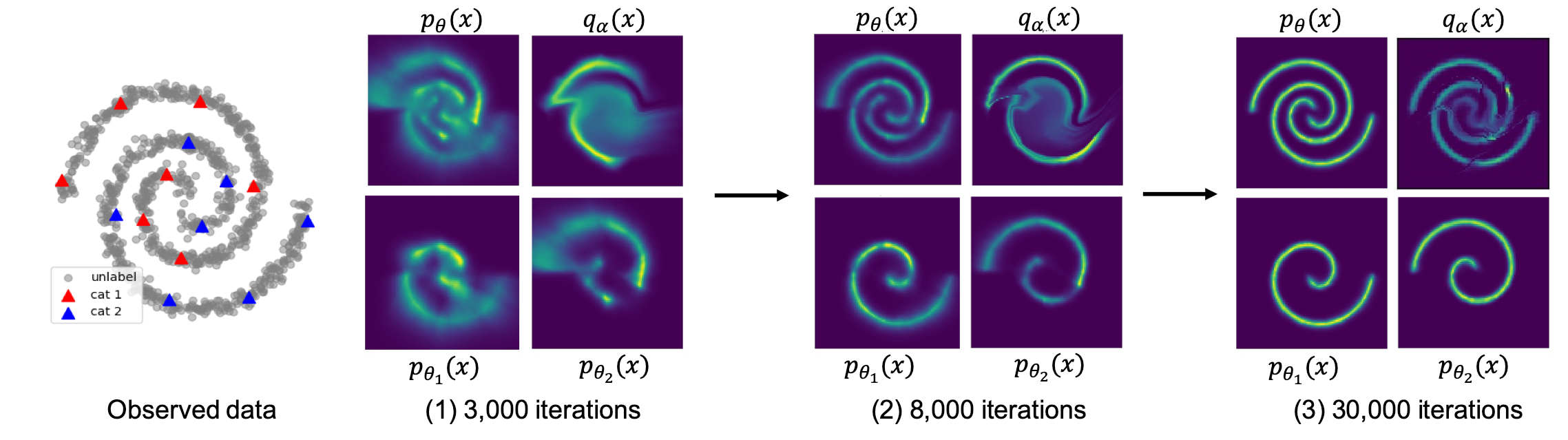}
    \captionof{figure}{Illustration of FCE for semi-supervised learning on a 2D example, where the data distribution is two spirals belonging to two categories. Within each panel, the top left is the learned unconditional EBM. The top right is the learned Glow model. The bottom are two class-conditional EBMs. For observed data, seven labeled points are provided for each category.}
    \label{fig: semi}
\end{center}
\end{figure*} 
In section \ref{sect: semi} we show that FCE can be generalized to perform semi-supervised learning. We emphasize that for semi-supervised learning, FCE not only learns a classification boundary or a posterior label distribution $p(y|x)$. Instead, the algorithm ends up with $K$ estimated probabilistic distributions $p(x|y=k), k = 1,... K$ for observed examples belonging to $K$ categories. Figure \ref{fig: semi} illustrates this point by showing the learning process on a 2D example, where the data distribution consists of two twisted spirals belonging to two categories. Seven labeled points are provided for each category. As the training goes, the unconditional EBM $p_\theta(x)$ learns to capture all the modes of the data distribution, which is in the form of a mixture of class-conditional EBMs $p_{
\theta_1}(x)$ and $p_{\theta_2}(x)$.  Meanwhile, by maximizing the objective function $L_{\rm label}(\theta)$ (eqn. \ref{eqn: label}), $p_\theta(x)$ is forced to project the learned modes into different spaces, resulting in two well-separated class-conditional EBMs. As shown in Figure \ref{fig: semi}, within a single mode of one category, the EBM tends to learn a smoothly connected cluster, which is often what we desire in semi-supervised learning. 

Then we test the proposed method on a dataset of real images. Following the setting in \cite{miyato2018virtual}, we use two types of CNN structures (\lq Conv-small\rq  and \lq Conv-large\rq) for EBMs, which are commonly used in state-of-the-art semi-supervised learning methods. See Appendix for detailed model structures. We start FCE from a pre-trained Glow model. Before the joint training starts, EBMs are firstly trained for $50,000$ iterations with the Glow model fixed. In practice, this helps EBMs keep pace with the pre-trained Glow model, and equips EBMs with reasonable classification ability. We report the performance at this stage as \lq FCE-init'. Also, since virtual adversarial training (VAT) \cite{miyato2018virtual} has been demonstrated as an effective regularization method for semi-supervised learning, we consider adopting it as an additional loss for learning the EBMs. More specifically, the loss is defined as the robustness of the conditional label distribution around each input data point against local purturbation. \lq FCE + VAT' indicates the training with VAT.

Table \ref{tabl: semi} summarizes the results of semi-supervised learning on SVHN dataset. We report the mean error rates and standard deviations over three runs. All the methods listed in the table belong to the family of semi-supervised learning methods. Our method achieve competitive performance to these state-of-the-art methods. \lq FCE + VAT' results show that the effectiveness of FCE does not overlap much with existing semi-supervised method, and thus they can be combined to further boost the performance. 
\begin{table}
\centering
 \caption{Semi-supervised classification error (\%) on the SVHN test set. $^\dagger$ indicates that we derive the results by running the released code. $^*$ indicates that the method uses data augmentation. The other cited results are provided by the original papers. Our results are averaged over three runs. }  
\footnotesize
 \begin{tabular}{lrr}
    \toprule
     \multirow{2}{*}{Method}   & \multicolumn{2}{c}{\# of labeled data} \\
     & \multicolumn{1}{c}{$500$} & \multicolumn{1}{c}{$1000$}\\
        \midrule
     SWWAE \cite{zhao2015stacked}       & & 23.56\\
     Skip DGM \cite{maaloe2016auxiliary} && 16.61 $(\pm 0.24)$ \\
     Auxiliary DGM \cite{maaloe2016auxiliary} &   & 22.86\\
     GAN with FM \cite{salimans2016improved} & 18.44 $(\pm 4.8)$ & 8.11 $(\pm 1.3)$\\
     VAT-Conv-small \cite{miyato2018virtual}  & & 6.83 $(\pm 0.24)$ \\
     \midrule
     \multicolumn{2}{l}{on Conv-small used in \cite{salimans2016improved, miyato2018virtual}}\\
     FCE-init & 9.42 $(\pm 0.24)$ &  8.50 $(\pm 0.26)$\\
     FCE & {\bf 7.05} $(\pm 0.28)$ & {\bf 6.35} $(\pm 0.12)$\\
     \midrule
     $\Pi$ model \cite{laine2016temporal} &7.05 $(\pm 0.30)$& 5.43 $(\pm 0.25)$\\
      VAT-Conv-large \cite{miyato2018virtual} & $^\dagger$8.98 $(\pm 0.26)$ & 5.77 $(\pm 0.32)$\\
           Mean Teacher \cite{tarvainen2017mean} & 5.45 $(\pm 0.14)$ & 5.21 $(\pm 0.21)$ \\
     $\Pi$ model$^*$ \cite{laine2016temporal} & 6.83 $(\pm 0.66)$& 4.95 $(\pm 0.26)$\\
     Temporal ensembling$^*$ \cite{laine2016temporal} & 5.12 $(\pm 0.13)$ & 4.42 $(\pm 0.16)$ \\ 
      \midrule
     \multicolumn{2}{l}{on Conv-large used in \cite{laine2016temporal, miyato2018virtual}}\\
      FCE-init & 8.86 $(\pm 0.26)$ &  7.60 $(\pm 0.23)$\\
     FCE & { 6.86} $(\pm 0.18)$  & 5.54 $(\pm 0.18)$\\
      FCE + VAT & {\bf 4.47} $(\pm 0.23)$  & {\bf 3.87} $(\pm 0.14)$\\ 
        \bottomrule
    \end{tabular}
    \label{tabl: semi}
\end{table}

\section{Conclusion}
This paper explores joint training of an energy-based model with a flow-based model, by combining the representational flexibility of the energy-based model and the computational tractability of the flow-based model.
We may consider the learned energy-based model as the learned representation, while the learned flow-based model as the learned computation.
This method can be considered as an adaptive version of noise contrastive estimation where the noise is transformed by a flow model to make its distribution closer to the data distribution and to make it a stronger contrast to the energy-based model. Meanwhile, the flow-based model is updated adaptively through the learning process, under the same adversarial value function. 

In future work, we intend to generalize the joint training method by combining the energy-based model with other normalized probabilistic models, such as auto-regressive models. We also intend to explore other joint training methods such as those based on adversarial contrastive divergence \cite{kim2016deep, dai2017calibrating, han2018divergence}  or divergence triangle \cite{han2018divergence}.

\subsubsection*{Acknowledgments}

The work is partially supported by DARPA XAI project N66001-17-2-4029 and ARO project W911NF1810296. We thank Pavel Sountsov, Alex Alemi, Matthew D. Hoffman and Srinivas Vasudevan for their helpful discussions. 

{\small
\bibliographystyle{ieee_fullname}
\bibliography{egbib}
}
\newpage
\appendix
\section{Model architectures}	
Table \ref{tab: ebm-arch} summarizes the EBM architectures used in unsupervised learning (subsections 4.1-4.3). The slope of all leaky ReLU (lReLU) \cite{maas2013rectifier} functions are set to $0.2$. For semi-supervised learning from a 2D example (subsection 4.4), we use the same EBM structure as the one used in unsupervised learning from 2D examples, except that for the top fully connect layer, we change the number of output channels to $2$, to model EBMs of two categories respectively. Table \ref{tab: ebm-semi-arch} summarizes the EBM architectures used in semi-supervised learning from SVHN (subsection 4.4).  After each convolutional layer, a weight normalization \cite{salimans2016weight} layer and a leaky ReLU layer is added. The slope of leaky ReLU functions is set to $0.2$. A weight normalization layer is added after the top fully connected layer. 

\begin{table}[h]
	\centering
	\caption{EBM architectures used in unsupervised learning}
	\label{tab: ebm-arch}
	\begin{tabular}{ll}
		\toprule
		2D data & SVHN / CIFAR-10  \\
		\midrule 
		fc. $128$ lReLU & $4\times4$ conv. $64$ lReLU, stride $2$  \\
		fc. $128$ lReLU & $4\times4$ conv. $128$ lReLU, stride $2$\\
		fc. $128$ lReLU & $4\times4$ conv. $256$ lReLU, stride $2$\\
		fc. $1$ & $4 \times 4$ conv. $1$, stride $1$ \\
		\bottomrule
	\end{tabular}
\end{table}

\begin{table}[h]
	\centering
	\caption{EBM architectures used in semi-supervised learning from SVHN}
	\label{tab: ebm-semi-arch}
	\begin{tabular}{cc}
		\toprule
		Conv-small & Conv-large  \\
		\midrule 
		\multicolumn{2}{c}{dropout, $p=0.2$}\\
		\midrule
		$3\times3$ conv. $64$, stride $1$ & $3\times3$ conv. $128$, stride $1$  \\
		$3\times3$ conv. $64$, stride $1$ & $3\times3$ conv. $128$, stride $1$  \\
		$3\times3$ conv. $64$, stride $2$ & $3\times3$ conv. $128$, stride $2$  \\
		\midrule
		\multicolumn{2}{c}{dropout, $p=0.5$}\\
		\midrule
		$3\times3$ conv. $128$, stride $1$ & $3\times3$ conv. $256$, stride $1$  \\
		$3\times3$ conv. $128$, stride $1$ & $3\times3$ conv. $256$, stride $1$  \\
		$3\times3$ conv. $128$, stride $2$ & $3\times3$ conv. $256$, stride $2$  \\
		\midrule
		\multicolumn{2}{c}{dropout, $p=0.5$}\\
		\midrule
		$3\times3$ conv. $128$, stride $1$ & $3\times3$ conv. $512$, stride $1$  \\
		$1\times1$ conv. $128$, stride $1$ & $1\times1$ conv. $256$, stride $1$  \\
		$1\times1$ conv. $128$, stride $1$ & $1\times1$ conv. $128$, stride $1$  \\
		\midrule
		\multicolumn{2}{c}{global max pool, $6\times6 \rightarrow 1\times1$}\\
		\multicolumn{2}{c}{fc. $128 \rightarrow 10$}\\
		\bottomrule
	\end{tabular}
\end{table}

For Glow model, we follow the setting of \cite{kingma2018Glow}. The architecture has multi-scales with levels $L$. Within each level, there are $K$ flow blocks. Each block has three convolutional layers (or fully-connected layers) with a width of $W$ channels. After the first two layers, a ReLU activation is added. Table \ref{tab: glow-arch} summarizes the hyperparameters for different datasets. 

\begin{table*}[h]
	\centering
	\caption{Hyperparameters for Glow model architectures}
	\label{tab: glow-arch}
	\begin{tabular}{lccccc}
		\toprule
		Dataset & Levels $L$ & Blocks per level $K$ & Width $W$ & Layer type & Coupling\\
		\midrule 
		2D data & 1 & 10 & 128 & fc & affine \\
		SVHN & 3 & 8 & 512 & conv & additive \\
		CelebA & 3 & 16 & 512 & conv & additive \\
		CIFAR-10 & 3 & 32 & 512 & conv & additive \\
		\bottomrule
	\end{tabular}
\end{table*}

\section{Synthesis comparison}
\begin{figure}[h]
\centering
	\includegraphics[width=.22\textwidth]{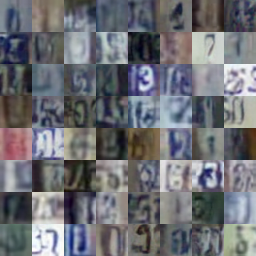}
	\hspace{0.1cm}
	\includegraphics[width=.22\textwidth]{fig/svhn_jsd_cut.png}
	\caption{Synthesized examples from Glow models learned from SVHN. Left panel is by MLE. Right panel is by our FCE. }
	\label{fig: svhn_compare}
\end{figure}

\begin{figure}[h]
\centering
	\includegraphics[width=.22\textwidth]{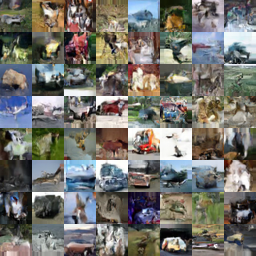}
	\hspace{0.1cm}
	\includegraphics[width=.22\textwidth]{fig/cifar_jsd_cut.png}
	\caption{Synthesized examples from Glow models learned from CIFAR-10. Left panel is by MLE. Right panel is by our FCE. }
	\label{fig: cifar_compare}
\end{figure}

\begin{figure}[h]
\centering
	\includegraphics[width=.22\textwidth]{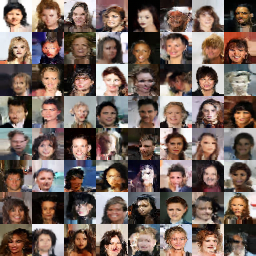}
	\hspace{0.1cm}
	\includegraphics[width=.22\textwidth]{fig/celeba_jsd.png}
	\caption{Synthesized examples from Glow models learned from CelebA. Left panel is by MLE. Right panel is by our FCE. }
	\label{fig: celeba_compare}
\end{figure}

In figures \ref{fig: svhn_compare}, \ref{fig: cifar_compare} and \ref{fig: celeba_compare}, we display the synthesized examples from Glow trained by MLE and our FCE.  

\end{document}